\documentclass[a4paper,twoside]{article}

\usepackage{epsfig}
\usepackage{subcaption}
\usepackage{calc}
\usepackage{amssymb}
\usepackage{amstext}
\usepackage{amsmath}
\usepackage{amsthm}
\usepackage{multicol}
\usepackage{pslatex}
\usepackage{apalike}
\usepackage{algorithm2e}
\usepackage[bottom]{footmisc}
\usepackage{hyperref}
\usepackage{SCITEPRESS}     % Please add other packages that you may need BEFORE the SCITEPRESS.sty package.

\begin{document}

\title{Reinforcement Learning-Enhanced Procedural Generation for Dynamic Narrative-Driven AR Experiences\\
\small DOI: \href{https://www.scitepress.org/PublicationsDetail.aspx?ID=LfPv9Lfiya8=&t=1}{10.5220/0013373200003912}
}

\author{\authorname{Aniruddha Srinivas Joshi\sup{1}}
\affiliation{\sup{1}Independent Researcher, M.Sc. in Games and Playable Media, University of California, Santa Cruz}
\email{ansjoshi@ucsc.edu}
\affiliation{\small \textnormal{\textbf{Cite as:} Aniruddha Srinivas Joshi, "Reinforcement Learning-Enhanced Procedural Generation for Dynamic Narrative-Driven AR Experiences," in \textit{Proceedings of the 20th International Joint Conference on Computer Vision, Imaging and Computer Graphics Theory and Applications - GRAPP}, 2025, pp. 385-397. SciTePress. DOI: \href{https://www.scitepress.org/PublicationsDetail.aspx?ID=LfPv9Lfiya8=&t=1}{10.5220/0013373200003912}.}}
}


\abstract{Procedural Content Generation (PCG) is widely used to create scalable and diverse environments in games. However, existing methods, such as the Wave Function Collapse (WFC) algorithm, are often limited to static scenarios and lack the adaptability required for dynamic, narrative-driven applications, particularly in augmented reality (AR) games. This paper presents a reinforcement learning-enhanced WFC framework designed for mobile AR environments. By integrating environment-specific rules and dynamic tile weight adjustments informed by reinforcement learning (RL), the proposed method generates maps that are both contextually coherent and responsive to gameplay needs. Comparative evaluations and user studies demonstrate that the framework achieves superior map quality and delivers immersive experiences, making it well-suited for narrative-driven AR games. Additionally, the method holds promise for broader applications in education, simulation training, and immersive extended reality (XR) experiences, where dynamic and adaptive environments are critical.}

\onecolumn \maketitle \normalsize \setcounter{footnote}{0} \vfill

\section{\uppercase{Introduction}}
\label{sec:introduction}

Procedural generation has become a cornerstone in the creation of diverse and scalable environments for games, enabling automated generation of complex layouts with minimal manual intervention. Although widely used in traditional gaming, its application in augmented reality (AR) remains limited, particularly in scenarios where environments need to dynamically adapt to gameplay narratives or physical surroundings. The Wave Function Collapse (WFC) algorithm \cite{Gumin16}, known for generating cohesive layouts through adjacency constraints, has been effective in creating static maps. However, it does not inherently address the challenges posed by narrative-driven experiences, where maps must align with evolving storylines and diverse contextual needs.

In this work, we extend the WFC algorithm to better serve the needs of narrative-driven AR games. By introducing environment-specific rules, our method tailors map generation to diverse settings, such as urban grids, open spaces, and dense terrains. These rules govern the placement of paths and features, ensuring maps are both visually coherent and thematically appropriate. To enhance adaptability, reinforcement learning (RL) refines generation decisions dynamically, adapting layouts to diverse gameplay requirements. Additionally, AR-specific features support real-time interactivity, enabling users to dynamically adjust maps to evolving gameplay narratives.

This approach bridges the gap between static procedural generation and the dynamic needs of narrative-driven AR games. By combining algorithmic enhancements with tools for real-time modification, our method delivers adaptive environments that enhance storytelling and gameplay. Beyond games, the framework can be applied to AR educational tools, simulation training, and other immersive experiences, offering a novel and practical advancement in state-of-the-art procedural content generation (PCG) methods.

To this end, this study addresses two primary research questions. Firstly, it evaluates how RL-enhanced WFC compares to traditional PCG methods in supporting the needs of narrative-driven augmented reality experiences. Secondly, it examines how the proposed method improves user experience by enabling dynamic, coherent, and immersive environments in narrative-driven AR games.

\section{\uppercase{Related Works}}

In the realm of interactive environments, procedural generation and machine learning (ML) have emerged as transformative technologies that enable the creation of dynamic and richly detailed content. This section delves into the evolution of these technologies from their foundational use in game development to their sophisticated integration within augmented reality systems. We examine how traditional procedural generation methods have been enhanced by learning algorithms to address the complexities of modern applications. The discussion underscores the need for more adaptable and context-aware generation methods, especially for enhancing user experiences in AR, virtual reality (VR), and extended reality (XR) environments, setting the stage for our proposed method designed to address these critical challenges.

\subsection{Procedural Generation in Games}

Procedural generation techniques have long been foundational for creating diverse and scalable content in games. One of the earliest techniques, Binary Space Partitioning (BSP), introduced a method to recursively divide space into convex sets, enabling efficient rendering and collision detection. Originally designed to solve the hidden surface problem \cite{fuchs1980}, BSP has since been adapted for generating structured layouts, such as dungeon levels, in modern games. Noise-based methods like Perlin Noise \cite{perlin1985image} and Simplex Noise improve naturalistic terrain generation, with Simplex Noise addressing computational inefficiencies and reducing artifacts. Cellular Automata \cite{johnson2010cellular} is another widely used approach for simulating organic structures such as caves or forests, evolving systems over time.

The Wave Function Collapse algorithm \cite{Gumin16} builds on earlier techniques for tile-based map generation. Notably, it shares significant similarities with the Model Synthesis algorithm \cite{merrell2011}. Model Synthesis differs in its approach to cell selection and its ability to modify the model in smaller blocks, which enhances its performance for generating larger and more complex outputs. A comparative analysis highlights their conceptual overlap and differences in implementation \cite{merrell2021comparison}. Unlike earlier techniques, WFC excels in maintaining structural coherence. However, it remains static in nature and lacks the ability to adapt dynamically to gameplay or narrative contexts, highlighting the need for more flexible and context-aware procedural generation methods, especially for interactive applications like augmented reality.

\subsection{Machine learning in Procedural Generation}

Machine learning has greatly expanded the possibilities of procedural content generation by enabling systems to adaptively generate content based on learned patterns. Methods like Generative Adversarial Networks (GANs) and Variational Autoencoders (VAEs) \cite{liu2021} are commonly used to create high-quality game assets, including levels and textures. These approaches introduce flexibility and adaptability, enhancing traditional PCG techniques.

RL has also shown promise for procedural tasks requiring sequential decision-making. For instance, RL-based frameworks \cite{khalifa2020} demonstrate how RL agents can generate game levels by framing level design as a Markov Decision Process (MDP). Similarly, recent work highlights RL’s ability to balance quality, diversity, and playability in level generation \cite{nam2024}. Despite these advancements, ML-based methods are often applied to 2D or platformer games and have yet to be fully integrated into augmented reality or interactive 3D environments.

\subsection{Procedural Generation in Augmented Reality}

Procedural Content Generation has also been applied in augmented reality to enhance user interaction by dynamically adapting virtual content to physical environments. Recent work presents a pipeline for integrating pre-existing 3D scenes into AR environments, minimizing manual adjustments and ensuring alignment with physical spaces \cite{caetano2022}. Similarly, PCG has been used to tailor AR game levels to the player’s surroundings, dynamically adjusting elements like layout and difficulty to leverage physical affordances \cite{azad2021}.

While these studies illustrate the potential of PCG in AR, they often focus on predefined or static content and rarely explore dynamic procedural generation tailored to narrative-driven gameplay. This work addresses these gaps by introducing adaptive PCG techniques for dynamically generating AR maps aligned with both narrative and gameplay needs, enabling real-time interactivity and customization.

\subsection{AR/VR/XR in Narrative Games}

AR, VR, and XR technologies have increasingly been used to create immersive environments for narrative-driven games. Research demonstrates how spatial interactivity can enhance storytelling by embedding narratives into physical spaces, providing players with unique, location-aware experiences \cite{viana2014}. Similarly, mobile AR studies examine the challenges of balancing user freedom with narrative control, highlighting AR’s potential for supporting interactive storytelling \cite{nam2015}.

Although these works showcase AR and XR’s strengths in narrative gaming, they often rely on manually designed environments, limiting scalability and adaptability. Few approaches incorporate procedural generation to dynamically align narratives with generated virtual spaces. This work builds on these foundations by integrating PCG into AR-specific features, enabling the dynamic creation of interactive environments that evolve alongside narrative-driven gameplay.

\section{\uppercase{Method}}
\label{sec:method}

In this section, we elucidate the proposed approach, which integrates reinforcement learning with the WFC algorithm \cite{Gumin16} to procedurally generate grid-based, immersive 3D maps in augmented reality for narrative-driven games. The RL-enhanced WFC method builds on the foundational WFC algorithm, originally designed for tile-based map generation using adjacency constraints. Our approach incorporates biome-specific constraints and reinforcement learning to optimize the generation process, ensuring biome coherence and enhancing path layouts.

We focus on three distinct biomes, each with unique layout and art styles:

\begin{itemize}
    \item \textbf{City}: A structured environment with continuous paths including pathways and buildings. Designed for interconnected urban settings that facilitate navigation.
    \item \textbf{Desert}: A sparse environment characterized by open areas and minimal impassable tiles such as boulders and cacti.
    \item \textbf{Forest}: A natural setting featuring paths between dense obstacles like trees and rocks, interspersed with open clearings.
\end{itemize}

\begin{figure}[!h]
    \centering
    \includegraphics[width=0.45\textwidth]{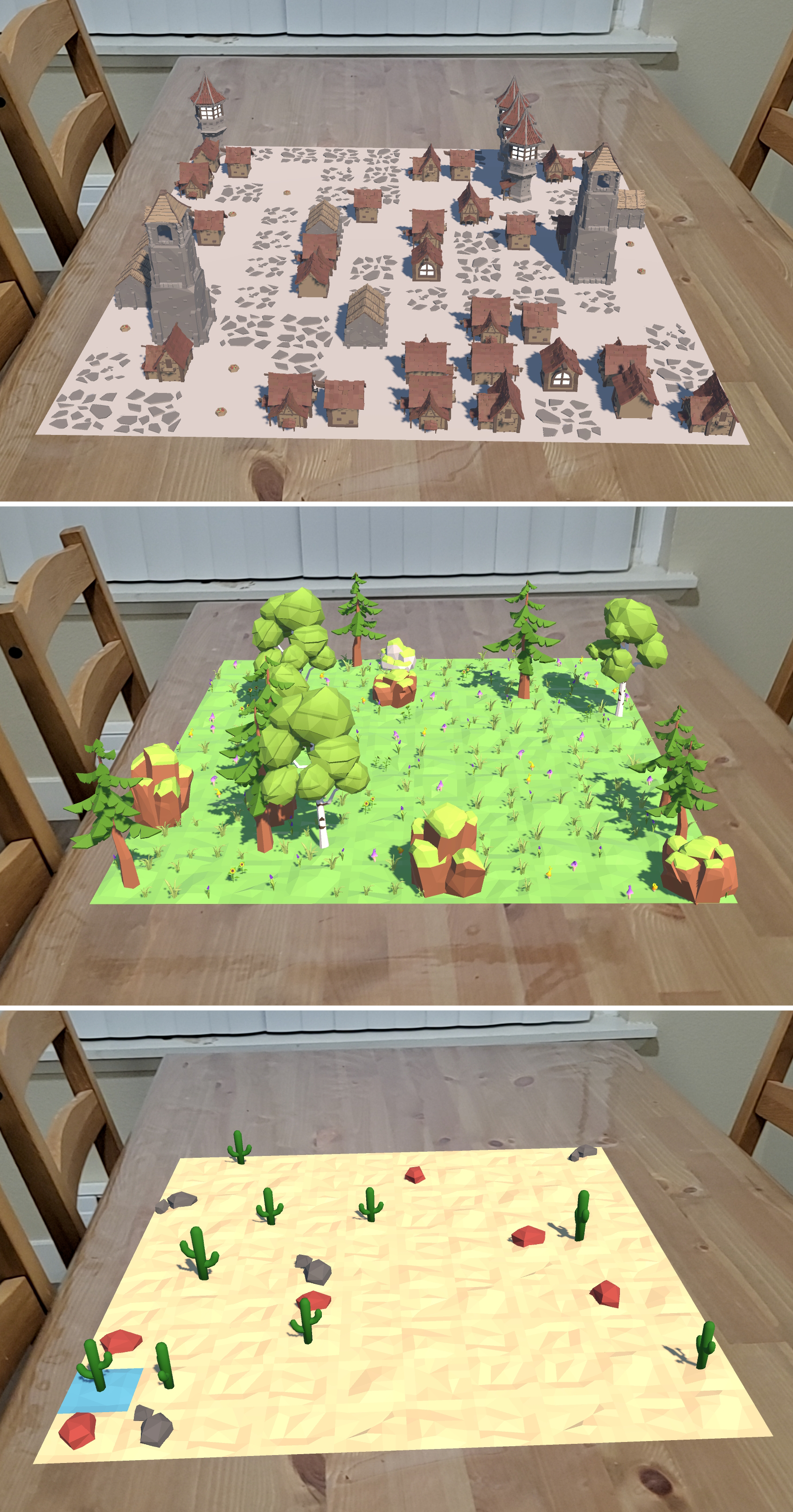}
    \caption{\emph{Real-time screen capture of 10×10 grids for City, Desert, and Forest biomes.}}
    \label{fig:biomes_merged}
\end{figure}

Figure~\ref{fig:biomes_merged} shows examples of each biome generated as a 10×10 grid tabletop map in AR. These biome-specific configurations influence the RL-enhanced WFC algorithm by defining tile types, adjacency rules, and path continuity, allowing the generated terrain to align with the intended narrative and gameplay.

We have designed our approach to generate a map in Augmented Reality for the narrative game \textit{Dungeons and Dragons} (D\&D) \cite{dnd2014}. The method includes interactive controls that enable the Dungeon Master (DM) to modify the generated AR map in real time. This enhances gameplay by allowing adjustments that align closely with the evolving narrative.

\subsection{Proposed Procedural Generation Method}
\label{wfc_implementation}

We now present the proposed procedural generation method aimed at constructing dynamic and interactive environments. Fundamental to our approach are the concepts of 'cell' and 'tile'. A cell is the basic unit of the grid capable of assuming multiple potential states known as tile options. A tile represents one of these options, each defined by unique characteristics such as type, adjacency rules, and visual representations (including 3D models). When a cell has a specific tile option selected, it is said to be 'collapsed' reducing its potential states to that single option. Conversely, a cell with multiple possible tile options remains 'non-collapsed' allowing for further decision-making as the algorithm progresses.

Our method employs the following key input parameters:
\begin{enumerate}
    \item Grid dimensions specifying the length and breadth of the grid.
    \item An array of all possible tiles, each detailed with properties such as adjacency rules, tile type, and the corresponding 3D model for rendering.
    \item A dynamic term from the RL agent ($r_{\text{RL}}$) that adjusts tile weight calculations to optimize procedural generation based on gameplay dynamics and environmental conditions. More details on this are discussed in Section~\ref{rl_method}.
\end{enumerate}

A high-level diagram of the proposed method is presented in Figure~\ref{fig:method_diagram}. The method implements the following key steps:

\begin{figure}[!h]
    \centering
    \includegraphics[width=\linewidth]{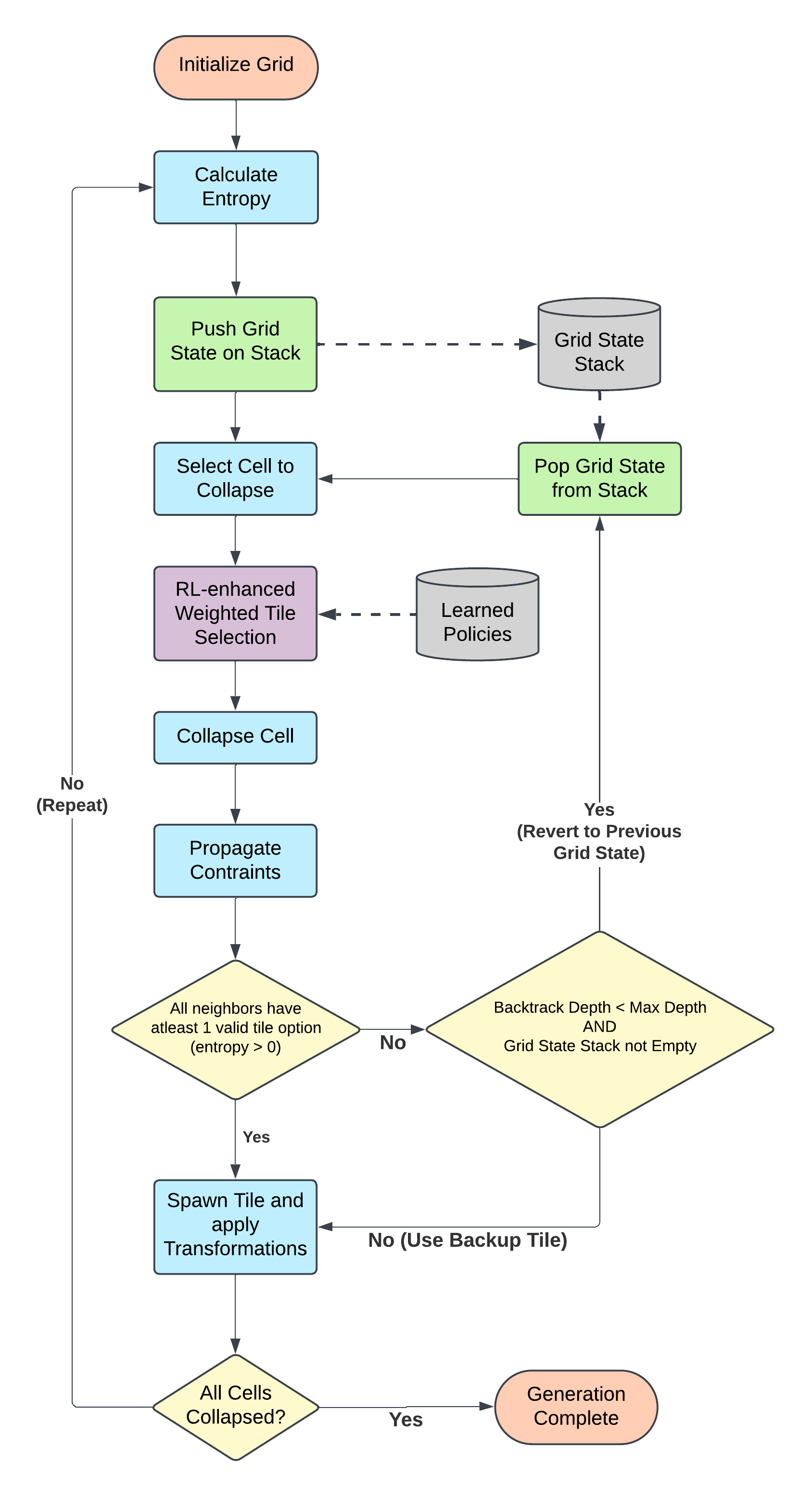}
    \caption{\emph{High-level diagram of the proposed method.}}
    \label{fig:method_diagram}
\end{figure}

\begin{enumerate}
    \item \textbf{Initialize Grid}: A grid of empty cells is generated based on the specified length and breadth dimensions. Each cell is initialized with all possible tile options according to the selected biome.
    \item \textbf{Calculate Entropy}: For each cell, the entropy is calculated as the count of valid tile options. The resulting grid state is pushed onto a stack.
    \item \textbf{Select Cell to Collapse}: Cells with lower entropy are prioritized for collapse. If multiple cells share the same entropy, one of these cells is picked at random (this always occurs in the first iteration where all cells have the same entropy).
    \item \textbf{Select Tile with RL-enchanced Weighted Randomness}: For the selected cell, a tile option is chosen based on weighted randomness. The calculation of these tile weights is detailed in Section~\ref{tile_weight_calculation}, while the integration of RL to refine tile selection is described in Section~\ref{rl_method}.
    \item \textbf{Collapse the Cell}: The selected cell is collapsed to its chosen tile, reducing its possible states to just that tile.
    \item \textbf{Propagate Constraints}: After the collapse, neighboring cells are updated to remove any tile options that conflict with the collapsed tile’s adjacency constraints. The neighborhood includes the cells directly up, down, left, and right of the collapsed cell, maintaining consistency with traditional WFC neighborhood rules. Path layout strategies as detailed in \ref{path_layout_strategies} are applied here to guide path connectivity.
    \item \textbf{Backtrack if Necessary}: While updating neighboring cells tile options, if any neighboring cell has no valid tile options remaining (i.e., entropy = 0), the algorithm reverts to a previous grid state saved on stack and attempts a different tile configuration. Backtracking is tracked using a depth counter, and if the depth reaches a predefined maximum limit (Max Depth) or the stack is empty, a default tile is used to resolve the deadlock and allow the algorithm to proceed.
    \item \textbf{Spawn and Apply Transformations}: The tile content is instantiated at the corresponding cell location. Transformations (rotation, symmetry, and scaling) are applied to the spawned tile to add visual variety while maintaining coherence.
    \item \textbf{Repeat Until Completion}: Steps 2 to 8 are repeated until every cell in the grid is collapsed, completing the map.
\end{enumerate}

\subsubsection{Tile Weight Calculation and Selection}
\label{tile_weight_calculation}
As we begin this section, it is crucial to acknowledge that Equation \ref{eq:tile_weight} has been extended to incorporate contributions from the RL agent. A detailed discussion on this topic will follow in Section \ref{rl_method}. Our current focus will be on the base implementation of weighted random selection.

Each cell in the grid has a list of possible tile options, and each tile option is assigned a weight based on how well it aligns with its neighboring cells. Tiles that fit better with neighbors are assigned higher weights to increase their likelihood of selection.

If a given cell has a total of \(T\) tile options, then the weight for the \(i^{\text{th}}\) tile option denoted \(w_i\) is calculated as follows:

\begin{equation}
\label{eq:tile_weight}
w_i = w_0 + \sum_{n \in \mathcal{N}} s_n
\end{equation}
where:
\begin{itemize}
    \item \( w_0 \) is the base weight (\( w_0 = 1.0 \)),
    \item \( \mathcal{N} \) represents the neighboring cells,
    \item \( s_n \) is the adjacency score for neighbor cell \( n \):
    \[
    s_n =
    \begin{cases}
        0, & \text{if } n \text{ is non-collapsed} \\
        & \textit{(i.e., n does not have a tile selected)}, \\
        1.5, & \text{if } n \text{ has a compatible tile} \\
        & \textit{(i.e., n's tile adheres to adjacency rules)}, \\
        0.5, & \text{if } n \text{ has an incompatible tile} \\
        & \textit{(i.e., n's tile violates adjacency rules)}
    \end{cases}
    \]

\end{itemize}

After calculating weights for \(T\) tile options, a weighted random selection is performed:
\begin{enumerate}
    \item Compute the total weight \( W = \sum_T w_i \) for all tile options.
    \item Generate a random number \( r \in [0, W] \).
    \item Select the tile with smallest index \( j \) such that \( \sum_{k=1}^j w_k \geq r \).
\end{enumerate}

This tile selection method ensures that tiles with higher weights (those that fit well with their neighbors) are more likely to be chosen, while still allowing some randomness in choice.

\subsubsection{Path Layout Strategies}
\label{path_layout_strategies}

The proposed method assumes two types of tiles:
\begin{enumerate}
    \item Path tiles: These are tiles that players can traverse (e.g., roads, grass).
    \item Impassable tiles: These are tiles that cannot be traversed and block movement (e.g., boulders, trees).
\end{enumerate}

Depending on the selected biome, the path layout can be continuous (e.g., city biome) or sparse (e.g., forest biome). These layouts are created by applying different path constraint strategies during the propagation step.

\paragraph{Continuous Path Layouts:} Strict adjacency constraints are enforced to ensure path connectivity. Once a path tile is selected, these constraints are enforced during propagation:

\begin{itemize}
    \item \textbf{Filtering Valid Tile Options}: For each neighboring cell, valid tile options are constrained to include only those that can connect to the path tile. This prioritization ensures that most neighboring cells favor path-compatible tiles, maintaining continuous connectivity.
    \item \textbf{Inclusion of Impassable Tiles}: Impassable tiles are only considered valid if they satisfy specific adjacency rules, such as requiring at least one adjacent path tile to preserve navigability. This approach allows for the integration of buildings, walls, or other impassable elements without disrupting the functionality of the map.
    \item \textbf{Weighted Randomness Favors Continuity of Path}: During tile selection, weighted randomness biases selection toward path tiles while allowing occasional placement of impassable tiles for diversity.
\end{itemize}

\paragraph{Sparse Path Layouts:} Relaxed adjacency constraints are used, allowing paths to be more scattered with impassable tiles (e.g., trees, rocks) interspersed among path tiles, creating a more open, fragmented layout. This is achieved through the following mechanisms:

\begin{itemize}
    \item \textbf{Random Application of Continuous Path Constraints}: When a path tile is placed, for each neighboring cell, there is a 50\% chance of applying continuous path constraints as previously described.
    \item \textbf{Flexible Tile Options}: If continuous path constraints are not applied, the neighboring cell retains a wider range of valid tile options, including path tiles and impassable tiles. This promotes more randomness and contributes to the sparse layout's fragmented structure.
    \item \textbf{Weighted Randomness Favors Variety}: In sparse layouts, weighted randomness slightly favors impassable tiles because they are more numerous than path tiles. This leads to more scattered obstructions and open spaces.
\end{itemize}

By introducing gaps in connectivity and balancing paths with impassable tiles, the resulting map achieves a more natural and unstructured appearance. This layout aligns with the aesthetics of open environments such as forests and deserts.

\subsection{Reinforcement Learning for Procedural Map Generation}
\label{rl_method}
RL is integrated to dynamically adjust tile weights in the WFC algorithm. By tailoring tile weights to biome-specific characteristics, this approach improves the coherence, completeness, and efficiency of map generation. The Proximal Policy Optimization (PPO) algorithm is chosen for its stability and ability to train effective policies while enabling moderate exploration \cite{schulman2017proximalpolicyoptimizationalgorithms}.

PPO uses a clipped surrogate objective function to stabilize policy updates:
\begin{equation}
    L^{CLIP}(\theta) = \mathbb{E}_t \left[\min\left(r_t(\theta)\hat{A}_t, \text{clip}(r_t(\theta), 1 - \epsilon, 1 + \epsilon)\hat{A}_t\right)\right]
\end{equation}
where:
\begin{itemize}
    \item \( \theta \) represents the policy network parameters
    \item \( t \) is the timestep
    \item \( r_t(\theta) \) is the probability ratio between new and old policies at \( t \)
    \item \( \hat{A}_t \) is the advantage estimate at \( t \), quantifying the relative benefit of actions
    \item \( \epsilon \) is the clipping threshold to limit policy update ratios
\end{itemize}

The PPO algorithm is used to train an RL agent to learn a policy that adjusts tile weights dynamically during map generation. Through episodic interactions with the WFC system, the agent observes the current grid state, biome type, and layout requirements to determine optimal adjustments. The policy learned by the agent is designed to maximize cumulative rewards, encouraging map generation that is efficient, complete, and biome-coherent.

Building on the baseline tile weight formula defined in Equation~\ref{eq:tile_weight}, the updated formula incorporates a dynamic adjustment term, \( r_{\text{RL}} \), derived from the RL agent's policy:
\begin{equation}
\label{eq:rl_tile_weight}
w_i = w_0 + \sum_{n \in \mathcal{N}} s_n + r_{\text{RL}}
\end{equation}
where \( r_{\text{RL}} \) accounts for real-time adjustments guided by the RL agent's policy.

This updated formula enables real-time tile weight adaptation, allowing the system to account for biome-specific variations while maintaining coherence and efficiency during map generation.

\subsubsection{Agent Training}
\label{method_agent_training}
Agent training proceeds in episodes, where each episode represents a single map generation task. At the beginning of each episode, the WFC system initializes an empty grid and the agent receives an observation. This observation encodes the current grid state as a binary representation of collapsed and uncollapsed cells, the biome type as a one-hot vector (e.g., Forest, City, or Desert), and the layout type as a binary value indicating sparse or continuous layouts.

The agent outputs a continuous adjustment value (\(RL\ Score\)), which dynamically modifies tile weights during map generation. The WFC algorithm uses these adjusted weights to generate a map, linking the agent’s actions to the resulting layout. After map generation, the agent evaluates the map’s quality using a reward function, which comprises three components:

\begin{enumerate}
    \item \textbf{Completeness:} Rewards fully collapsed grids and penalizes incomplete or invalid configurations. This is formally defined as:
    \begin{equation}
    C = 
    \begin{cases} 
    +1, & \text{if all cells are collapsed and valid,} \\
    -0.5, & \text{if some cells remain uncollapsed,} \\
    -1, & \text{if grid contains invalid configurations}
    \end{cases}
    \label{eq:completeness}
    \end{equation}

    \item \textbf{Biome Coherence:} Rewards alignment with biome-specific characteristics, such as sparse layouts for Forests, continuous paths for Cities, and open spaces for Deserts. It is calculated as:
    \begin{equation}
    B = \frac{\sum_{j=1}^{T} b_j}{T}
    \label{eq:biome_coherence}
    \end{equation}
    where \( b_j = 1 \) if the \( j^{\text{th}} \) tile adheres to biome-specific adjacency constraints, and \( T \) is the total number of tiles.

    \item \textbf{Efficiency:} Rewards faster map generation and penalizes retries or backtracking. Efficiency is defined as:
    \begin{equation}
    E = k_1 \cdot (S_{\text{max}} - S_{\text{used}}) - k_2 \cdot B
    \label{eq:efficiency}
    \end{equation}
    where:
    \begin{itemize}
        \item \(k_1 = 0.2\) is the reward factor for minimizing steps
        \item \(k_2 = 0.1\) is the penalty factor for backtracking
        \item \(S_{\text{max}}\) is the maximum allowed steps for map generation
        \item \(S_{\text{used}}\) is the number of steps taken to complete the grid
        \item \(B\) is the number of backtracking operations performed
    \end{itemize}
\end{enumerate}

The total reward for an episode is given by:
\begin{equation}
R = C + B + E
\label{eq:total_reward}
\end{equation}

The agent receives this cumulative reward, which guides its policy updates. PPO adjusts the policy parameters to maximize future rewards, reinforcing beneficial actions while discouraging suboptimal ones. The clipped objective ensures incremental updates, maintaining training stability and preventing large policy shifts. Over successive episodes, the agent refines its policy, learning to associate specific tile weight adjustments with desirable map characteristics. By the end of training, the policy generalizes effectively across biomes, enabling adaptability to diverse map generation scenarios.

\subsubsection{Policy Deployment}

After training, the policy learned by the RL agent is deployed in inference mode i.e. it applies the learned adjustments to make decisions without further updates. During deployment, the system uses the policy to dynamically adjust tile weights based on the biome type and grid state in real-time. These adjustments ensure that the generated maps are coherent, complete, and efficient, aligning with the specific requirements of each biome.

\subsection{Mobile AR Implementation}
\label{mobile_ar_implementation}
The Augmented Reality mobile app was developed using Unity’s AR Foundation API \cite{ar_foundation}, which provides built-in plane detection to identify horizontal surfaces, such as tabletops, as valid spaces for placing virtual elements. Detected surfaces are visually highlighted to indicate valid placement areas, offering clear feedback to the user. Once a surface is confirmed through a screen tap, a raycast is performed from the tap position to the plane surface. The raycast ensures accurate placement of the procedural grid map, which is generated based on the specified dimensions and biome, and then aligned with the detected surface for an immersive and interactive experience.

To ensure the method is optimized for mobile devices, grid sizes were limited to a maximum of 15×15. This constraint prevents excessive computational overhead while also aligning with the physical constraints of typical tabletop surfaces, ensuring that the generated maps remain practical and immersive for mobile AR. Low-poly assets were used to reduce rendering demands, preserving visual fidelity while ensuring efficient performance. Additionally, tile backtracking depth during map generation was restricted to minimize unnecessary computations, enhancing responsiveness.

By integrating robust AR placement mechanisms with mobile-specific optimizations, the framework delivers detailed, biome-specific maps while maintaining real-time interactivity. These strategies ensure that mobile AR applications meet the needs of narrative-driven gameplay while remaining efficient and responsive.

\subsection{Interactive Narrative Control}
\label{interactive_narrative_control}

The proposed method provides dynamic narrative controls for real-time customization of procedurally generated AR environments. These interactive controls give DMs the ability to adapt the map to align with the evolving narrative. This allows the environment to respond to key narrative events and unexpected player actions, bridging the gap between procedural generation and narrative-driven gameplay.

\begin{figure}[!h]
    \centering
    \includegraphics[width=0.45\textwidth]{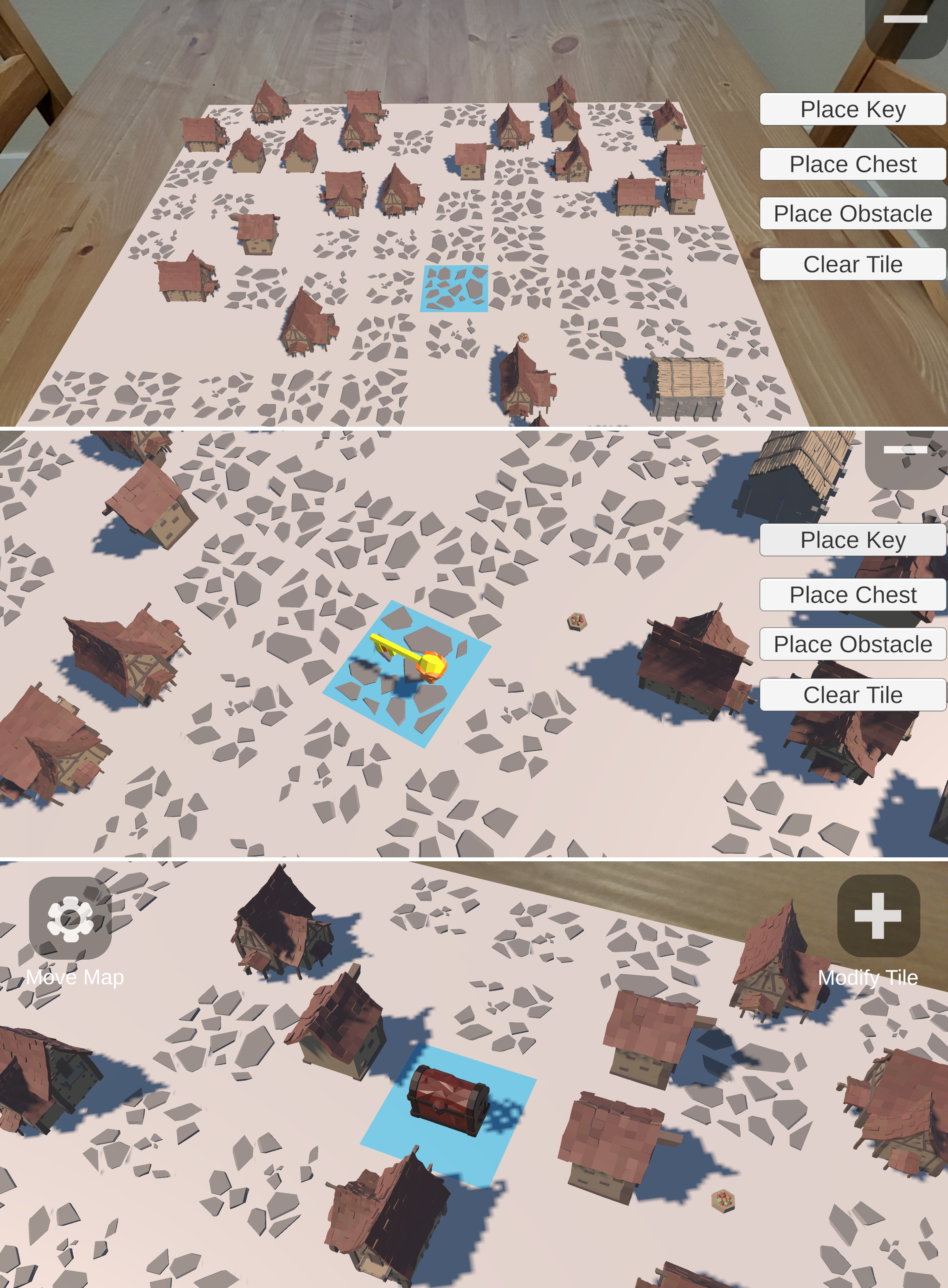}
    \caption{\emph{Real-time screen captures showcasing dynamic controls offered by the proposed method.}}
    \label{fig:interactive_controls}
\end{figure}

The key interactive features are:

\begin{enumerate}
    \item \textbf{Biome and Grid Size Selection}: The DM can select the biome (City, Desert, or Forest) and define grid dimensions at the start of map generation. This ensures the environment aligns with the story’s thematic and spatial requirements.
    
    \item \textbf{Clearing Paths}: Hidden routes can be revealed to guide players toward new objectives or advance the narrative. For instance, uncovering a passage after solving a puzzle ensures the story progresses naturally while encouraging exploration.
    
    \item \textbf{Blocking Paths}: Existing routes can be closed off to simulate story-driven events or introduce environmental challenges. Examples include blocking a path to represent a cave-in or using a locked gate to redirect players toward specific objectives.
    
    \item \textbf{Placing Special Objects}: Narrative-critical objects such as traps, treasure chests, keys, or locked doors can be dynamically added to the map. These objects embed story-specific interactions into the environment, such as requiring players to locate a key to progress or triggering traps during pivotal moments.
\end{enumerate}
The integration of these interactive controls into the procedural generation framework establishes a strong connection between the narrative and the environment. Unlike traditional PCG methods that produce static maps, the proposed framework enables real-time adaptability, transforming the environment from a passive backdrop into an active participant in the storytelling process. Figure~\ref{fig:interactive_controls} illustrates these dynamic controls.

By allowing DMs to clear paths, block routes, and place narrative-critical objects dynamically, the framework empowers them to align the environment with both predefined story arcs and emergent player-driven scenarios. For example, uncovering a hidden path can guide players toward key objectives, while introducing obstacles or rewards in response to player actions creates unique, unplanned narrative moments. This flexibility ensures that the environment remains engaging and contextually relevant.

Through this combination of real-time customization and procedural generation, the proposed method supports immersive, story-driven applications in AR gaming and other interactive environments. It demonstrates the potential of PCG to move beyond static content creation, enabling richer and more dynamic storytelling experiences.

\section{\uppercase{Experimental Evaluation}}

In this section, we present the experimental evaluation conducted to validate the effectiveness and efficiency of our proposed procedural generation method. The evaluation includes the training of the RL agent, a user study assessing interaction quality with AR applications, and a performance analysis comparing our method to baseline approaches. Together, these evaluations provide a comprehensive assessment of the proposed method’s application in generating procedural content for interactive AR environments.

\subsection{RL Agent Training}
\label{exp:rl_agent_training}

The RL agent was trained in Unity using ML-Agents with PyTorch for policy updates. Training was conducted on a Windows workstation equipped with an AMD Ryzen 9 5900X 12-Core processor and an NVIDIA RTX 4070 Super GPU. Unity's simulation environment handled procedural map generation during episodic interactions with the agent, providing the basis for policy learning.

The agent was trained using 15×15 grids, as this is the maximum grid size supported by the mobile application. Each biome (City, Forest, and Desert) was trained for 50,000 episodes, a value chosen to provide enough iterations for the policy to generalize across diverse map generation scenarios. 

Training parameters included a batch size of 64, a learning rate of \(3.0 \times 10^{-4}\), a buffer size of 2048, a discount factor (\(\gamma\)) of 0.99, and a clipping threshold (\(\epsilon\)) of 0.2. Observations consisted of biome-specific inputs and the grid state (collapsed/uncollapsed cells). A single continuous action \(r_{\text{RL}}\) dynamically adjusted tile weights during training.

The reward function as defined in Section~\ref{method_agent_training} guided the agent's policy updates by balancing Completeness, Biome Coherence, and Efficiency. For the Efficiency component, \(k_1 = 0.2\) and \(k_2 = 0.1\) were selected to encourage faster map generation while penalizing excessive backtracking. These values were chosen to balance the trade-offs observed during initial training runs. Higher backtracking penalties caused overly cautious behavior, while lower penalties made retries less impactful.

By the end of training, the agent effectively learned to adjust tile weights dynamically, enabling the generation of coherent, complete, and efficient maps across diverse biomes.

\subsection{User Study}

A user study was conducted with 28 participants with a diverse range of D\&D experience levels, as shown in Table \ref{tab:years_experience}. Each participant interacted with three AR applications, each implementing a different procedural generation algorithm:

\begin{itemize}
    \item \textbf{App 1:} Perlin Noise \cite{perlin1985image}
    \item \textbf{App 2:} Cellular Automata \cite{johnson2010cellular}
    \item \textbf{App 3:} Proposed Method
\end{itemize}

We conducted a statistical power analysis to determine the appropriate sample size for our within-subjects study. As with prior work in immersive media \cite{breves2021bridging} and \cite{hogberg2019gameful}, we assumed a medium effect size (\( f = 0.25 \)), a significance level of \( \alpha = 0.05 \), and a statistical power of \( 1 - \beta = 0.8 \) \cite{cohen2016power}. Using G* Power for a repeated measures ANOVA with three conditions, the required sample size was calculated to be 28 participants.

\begin{table}[h!]
\centering
\caption{Participant Distribution by Years of D\&D Experience}
\label{tab:years_experience}
\begin{tabular}{|c|c|}
\hline
Experience Level & Number of Participants \\ \hline
5+ years         & 6                       \\ \hline
3--5 years       & 6                       \\ \hline
1--3 years       & 7                       \\ \hline
Less than 1 year & 9                       \\ \hline
\end{tabular}
\end{table}

Participants rated the procedural generation algorithms on key dimensions using a 5-point Likert scale (1 = Poor, 5 = Excellent). The aggregated results for these dimensions are shown in Table~\ref{tab:user_response}. Questions asked assessed biome coherence, immersion, usability, visual quality, speed, and suitability, focusing on critical aspects of evaluating procedural content generation.

Following the presentation of the aggregated results in Table~\ref{tab:user_response}, it is important to note that while Likert scales are ordinal, the use of means for their analysis is aligned with common practices in Human-Computer interaction (HCI) research. This methodological choice, as supported by \cite{kaptein2010} helps in providing clear, actionable insights despite the ordinal nature of the data, especially useful in contexts with small sample sizes where robustness and ease of interpretation are prioritized.

Biome Coherence (Forest, Desert, City) evaluated the logical structure and realism of generated environments, supported by frameworks like \cite{togelius2011search} and \cite{shaker2016pcg}, which emphasize alignment with user expectations and natural archetypes to ensure quality and coherence. Biome Suitability evaluated alignment with biome-specific characteristics, such as sparse desert layouts or structured city grids, supported by frameworks for procedural generation and environmental coherence \cite{azad2021} and \cite{togelius2011search}. Immersion assessed user engagement, inspired by the Presence Questionnaire \cite{witmer1998presence}, while Generation Speed reflected usability principles from the System Usability Scale (SUS) \cite{brooke1995sus}. Visual Quality, adapted from the User Experience Questionnaire (UEQ) \cite{laugwitz2008ueq}, measured aesthetics, and Preferred App captured participants' overall impressions, commonly used in procedural content evaluations \cite{khalifa2020}. 

\begin{table}[h!]
\centering
\caption{User Study Ratings Across Procedural Generation Methods}
\label{tab:user_response}
\begin{tabular}{|l|c|c|c|}
\hline
\textbf{Metric}      & \textbf{App 1} & \textbf{App 2} & \textbf{App 3} \\ \hline
Forest Coherence     & 4.5            & 4.0            & 4.6            \\ \hline
Desert Coherence     & 3.8            & 3.3            & 4.4            \\ \hline
City Coherence       & 3.0            & 4.2            & 4.7            \\ \hline
Immersion            & 4.1            & 4.0            & 4.5            \\ \hline
Biome Suitability    & 3.7            & 3.8            & 4.6            \\ \hline
Generation Speed     & 4.7            & 4.3            & 3.6            \\ \hline
Visual Quality       & 4.2            & 4.0            & 4.7            \\ \hline
Preferred App        & 18\% (5)       & 14\% (4)       & 68\% (19)      \\ \hline
\end{tabular}
\end{table}

\begin{figure}[!h]
    \centering
    \includegraphics[width=0.45\textwidth]{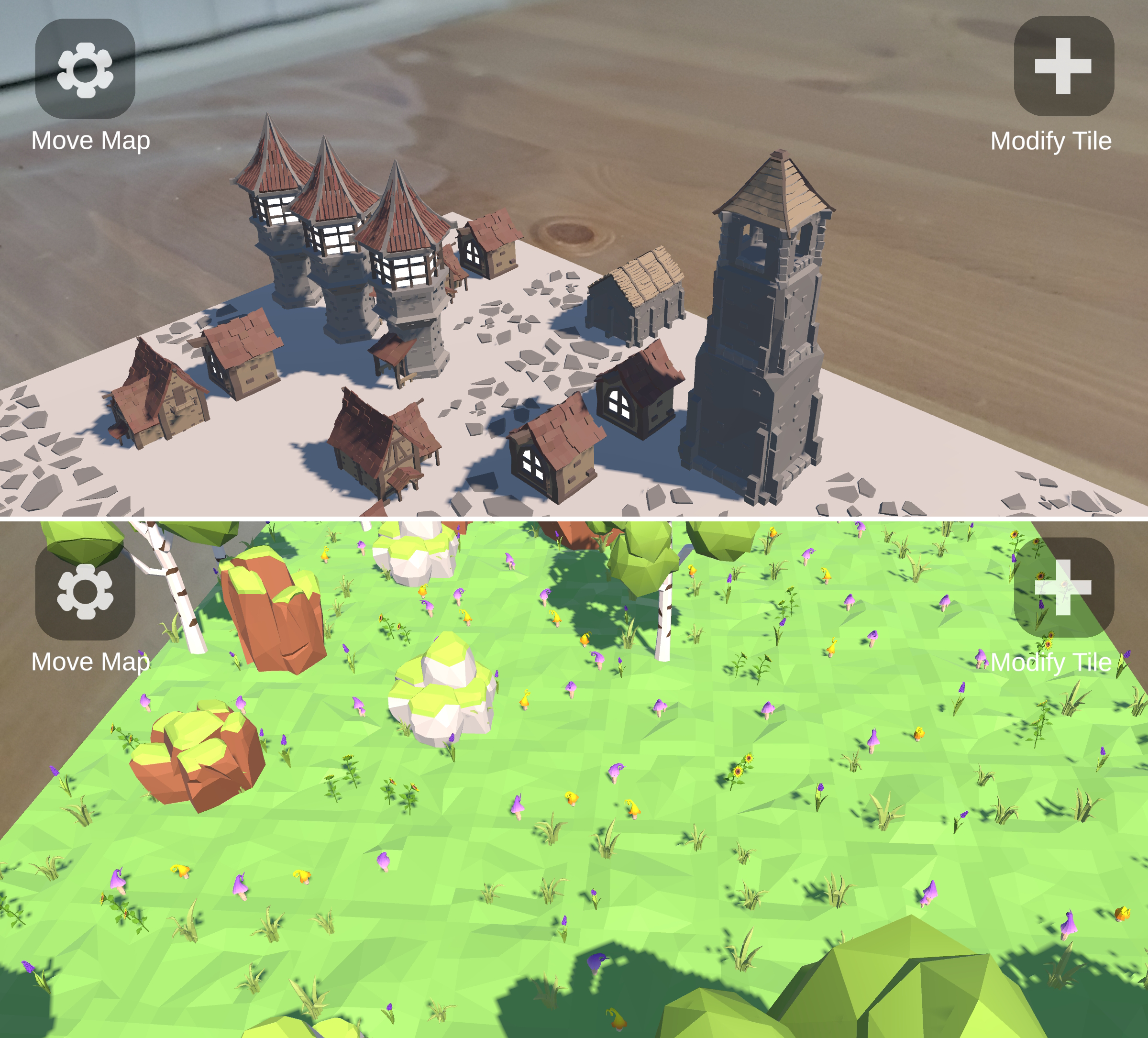}
    \caption{\emph{Close up real-time screen captures of environments generated by the proposed method.}}
    \label{fig:detailed_map_gen}
\end{figure}

The following points detail the key observations from Table~\ref{tab:user_response}:
\begin{itemize}
\item \textbf{Perlin Noise (App 1):}
    \begin{itemize}
        \item Strong performance in Forest coherence (4.5), offering natural visuals.
        \item Weak in City coherence (3.0) due to unrealistic structure.
    \end{itemize}

\item \textbf{Cellular Automata (App 2):}
    \begin{itemize}
        \item Performed well in City coherence (4.2).
        \item Lowest score in Desert coherence (3.3), struggling with sparse layouts.
    \end{itemize}

\item \textbf{Proposed Method (App 3):} 
    \begin{itemize}
        \item Rated highest for Forest coherence (4.6), Desert coherence (4.4), and City coherence (4.7).
        \item Participants praised biome-specific realism and coherence. Figure~\ref{fig:detailed_map_gen} supports this feedback with examples of generated environments.
        \item Slower generation speed (3.6) was noted as a drawback.
    \end{itemize}

\item \textbf{Overall:}  
    \begin{itemize}
        \item 68\% of participants preferred App 3, while 18\% chose App 1 and 14\% chose App 2.
    \end{itemize}
\end{itemize}

To evaluate the broader usability and suitability of the proposed method (App 3) for narrative games like \textit{Dungeons \& Dragons}, participants rated additional aspects beyond the key evaluation metrics using a 5-point Likert scale (1 = Poor, 5 = Excellent). Since the user interface remained identical across apps, questions on Ease of Use, Better than Traditional Map-Building Methods, Likelihood of Future Use, Satisfaction with Visual Quality, and Engagement in Customization were asked only once for the proposed method. The results of these ratings are summarized in Table~\ref{tab:usability_ratings}. These additional questions were informed by the System Usability Scale \cite{brooke1995sus} and the Presence Questionnaire \cite{witmer1998presence} to target broader usability and narrative-driven utility. This design ensured a focused evaluation of the algorithms, balancing comprehensiveness and usability to meet the study's objectives.

\begin{table}[ht]
\centering
\caption{Usability Ratings for the Proposed Method}
\label{tab:usability_ratings}
\begin{tabular}{|l|c|}
\hline
\textbf{Aspect}                                & \textbf{Rating (Mean)} \\ \hline
Ease of Use                                   & 4.1                   \\ \hline
Better than Traditional Methods & 4.0                   \\ \hline
Likelihood of Future Use                      & 4.0                   \\ \hline
Engagement in Customization                   & 3.8                   \\ \hline
\end{tabular}
\end{table}

\subsection{Computational Performance Analysis}
\label{performance_comparison}

We assess the computational performance of our proposed procedural generation method, focusing on its efficiency relative to the baseline techniques used in the user study. This evaluation provides insights into the method's practicality for implementation on mobile devices.

\subsubsection{Experimental Setup and Metrics}
The performance of the proposed method was evaluated alongside two baseline methods: Perlin Noise and Cellular Automata. The evaluation was conducted using a Samsung Galaxy S21 with a Snapdragon 888 processor and 8 GB of RAM.

Performance was evaluated by measuring the following key metrics:

\begin{enumerate}
    \item Time Taken: Total time required to generate the grid map.
    \item Minimum FPS: The lowest frame rate recorded during the generation process.
    \item Average FPS: The overall average frame rate maintained during map generation.
    \item FPS Recovery Time: Time taken for the system to stabilize back to 60 FPS after an FPS drop.
\end{enumerate}

Results were recorded for grid sizes of 5×5, 10×10, and 15×15 across each biome with each result averaged over 10 trials. These results are shown in Tables~\ref{tab:perlin_noise_performance}, \ref{tab:ca_performance}, and ~\ref{tab:proposed_method_performance}.

% Table for Perlin Noise metrics goes here (Table~\ref{tab:perlin_noise_performance})
\begin{table}[!t]
\centering
\caption{Computational Performance Metrics for Perlin Noise}
\label{tab:perlin_noise_performance}
\resizebox{\columnwidth}{!}{%
\begin{tabular}{|c|c|c|c|c|c|}
\hline
\textbf{Biome} & \textbf{Grid Size} & \textbf{Time (s)} & \textbf{Min FPS} & \textbf{Avg FPS} & \textbf{FPS Recovery (s)} \\ \hline
Forest         & 5×5 = 25          & 0.534                   & 51.25                & 54.3075          & 0.117                          \\ 
               & 10×10 = 100       & 1.763                   & 52.333               & 57.644           & 0.467                          \\ 
               & 15×15 = 225       & 3.841                   & 53.333               & 59.106           & 0.512                          \\ \hline
               %& 20×20 = 400       & 7.865                   & 40.75                & 55.14            & 0.613                          \\ \hline
Desert         & 5×5 = 25          & 0.518                   & 52.667               & 56.115           & 0.0833                         \\ 
               & 10×10 = 100       & 1.753                   & 54                   & 58.129           & 0.564                          \\ 
               & 15×15 = 225       & 3.888                   & 50.667               & 58.671           & 0.612                          \\ \hline
               %& 20×20 = 400       & 6.747                   & 54.333               & 56.252           & 0.467                          \\ \hline
City           & 5×5 = 25          & 0.478                   & 56                   & 58.859           & 0.028                          \\ 
               & 10×10 = 100       & 1.802                   & 51                   & 56.967           & 0.662                          \\ 
               & 15×15 = 225       & 3.859                   & 51.667               & 58.899           & 0.473                          \\ \hline
               %& 20×20 = 400       & 6.744                   & 54.333               & 59.585           & 0.45                           \\ \hline
\end{tabular}%
}
\end{table}

% Table for Cellular Automata metrics goes here (Table~\ref{tab:ca_performance})
\begin{table}[!t]
\centering
\caption{Computational Performance Metrics for Cellular Automata (CA)}
\label{tab:ca_performance}
\resizebox{\columnwidth}{!}{%
\begin{tabular}{|c|c|c|c|c|c|}
\hline
\textbf{Biome} & \textbf{Grid Size} & \textbf{Time (s)} & \textbf{Min FPS} & \textbf{Avg FPS} & \textbf{FPS Recovery (s)} \\ \hline
Forest         & 5×5 = 25          & 0.528                   & 51.667               & 55.333           & 0.321                          \\ 
               & 10×10 = 100       & 1.8052                  & 50.2                 & 56.818           & 0.5838                         \\ 
               & 15×15 = 225       & 4.07                    & 49                   & 56.894           & 0.523                          \\ \hline
               %& 20×20 = 400       & 8.903                   & 33.5                 & 48.802           & 0.905                          \\ \hline
Desert         & 5×5 = 25          & 0.517                   & 52.667               & 55.295           & 0.094                          \\ 
               & 10×10 = 100       & 1.835                   & 49.25                & 56.094           & 0.888                          \\ 
               & 15×15 = 225       & 3.88                    & 51.333               & 58.715           & 0.617                          \\ \hline
               %& 20×20 = 400       & 6.753                   & 63.667               & 59.528           & 0.483                          \\ \hline
City           & 5×5 = 25          & 0.589                   & 47.667               & 53.103           & 0.0444                         \\ 
               & 10×10 = 100       & 1.845                   & 47.333               & 56.02            & 0.611                          \\ 
               & 15×15 = 225       & 3.98                    & 46                   & 57.58            & 1.234                          \\ \hline
               %& 20×20 = 400       & 7.577                   & 43                   & 54.689           & 0.84                           \\ \hline
\end{tabular}%
}
\end{table}

\begin{table}[!t]
\centering
\caption{Computational Performance Metrics for Proposed Method}
\label{tab:proposed_method_performance}
\resizebox{\columnwidth}{!}{%
\begin{tabular}{|c|c|c|c|c|c|}
\hline
\textbf{Biome} & \textbf{Grid Size} & \textbf{Time (s)} & \textbf{Min FPS} & \textbf{Avg FPS} & \textbf{FPS Recovery (s)} \\ \hline
Forest         & 5×5 = 25          & 0.614             & 47.2             & 52.346           & 0.08                      \\
               & 10×10 = 100       & 4.332             & 16.4             & 29.6426          & 2.279                     \\
               & 15×15 = 225       & 24.0454           & 5.6              & 14.7964          & 18.0716                   \\ \hline
               %& 20×20 = 400       & 50.539            & 5                & 10.981           & 43.991                    \\ \hline
Desert         & 5×5 = 25          & 0.5906            & 48.6             & 52.962           & 0.0732                    \\
               & 10×10 = 100       & 4.189             & 16               & 31.099           & 2.0714                    \\
               & 15×15 = 225       & 21.059            & 5.833            & 17.522           & 15.371                    \\ \hline
               %& 20×20 = 400       & 45.568            & 5.4              & 13.016           & 39.137                    \\ \hline
City           & 5×5 = 25          & 0.709             & 43.75            & 49.991           & 0.075                     \\
               & 10×10 = 100       & 7.592             & 8.2              & 22.232           & 3.477                     \\
               & 15×15 = 225       & 29.98             & 3.5                & 12.175           & 29.61                     \\ \hline
               %& 20×20 = 400       & 84.15             & 3                & 8.237            & 73.781                    \\ \hline
\end{tabular}%
}
\end{table}

\subsubsection{Observations}
The following observations were made from the comparative results:
\begin{itemize}
    \item Perlin Noise was the fastest method across all biomes with Cellular Automata performing slightly slower and the proposed method requiring significantly more time, particularly for larger grid sizes.
    \item Performance consistently degraded with increasing grid size across all methods with longer generation times, lower minimum FPS, and prolonged FPS recovery times.
    \item Desert biome exhibited consistently better performance with shorter generation times and higher FPS across methods while the Forest and City biomes were more computationally demanding.
\end{itemize}

\subsection{Results}

The computational performance and user feedback evaluations reveal key trade-offs between quality and efficiency for the proposed method compared to Perlin Noise and Cellular Automata in mobile AR environments.

The evaluation of generation times as shown in Tables~\ref{tab:perlin_noise_performance}, \ref{tab:ca_performance} and \ref{tab:proposed_method_performance}, demonstrated that the proposed method required significantly longer times across all grid sizes and biomes. While slower, the proposed method produced highly detailed, biome-specific maps that were rated superior in user studies, reinforcing its value for applications where quality and coherence are critical.

Performance trends across grid sizes showed that larger grids led to longer generation times, reduced minimum FPS and increased FPS recovery times across all methods. The Desert biome consistently performed better with shorter generation times and higher FPS values while the City and Forest biomes posed greater computational challenges due to their complexity. Despite these challenges, the proposed method achieved significantly higher map coherence and quality across all biomes, particularly for more intricate layouts in the City environment.

The user study results validated the proposed method’s advantages, with participants consistently rating its maps as more immersive, visually coherent, and narrative-friendly than those from baseline methods. These findings highlight its suitability for narrative-driven AR applications where immersion and contextual alignment take precedence over speed. While WFC without RL was not explicitly evaluated in this study, future work could include such a comparison to further quantify the RL-enhanced framework’s contributions to coherence and adaptability. Moreover, the results confirm that the research questions posed in the Introduction were effectively addressed. Specifically, the proposed method supports narrative-driven AR applications by delivering immersive and coherent environments, while offering performance comparable to traditional PCG techniques in this context.

Although the proposed method's computational demands are higher, they remain acceptable for mobile AR applications, especially for narrative-driven scenarios prioritizing immersion over speed. Potential optimizations, such as asynchronous map generation in the background could mitigate FPS drops and reduce perceived delays. Device specifications also play a significant role, with lower-spec devices facing greater challenges when handling larger grids.

Overall, the findings from both the user study and computational evaluation demonstrate that the research questions posed at the beginning of this work were effectively addressed. The proposed method delivers immersive and coherent environments for narrative-driven AR applications, while achieving a balance between computational demands and the quality required for such contexts.

\section{\uppercase{Conclusions}}
\label{sec:conclusion}

In this work, a reinforcement learning-enhanced wave function collapse based procedural generation method was developed and evaluated for mobile AR environments. The proposed method demonstrated superior coherence and map quality, making it particularly well-suited for narrative-driven games where immersion and environmental detail are critical. While the method introduces higher computational demands compared to baseline approaches, user studies consistently rated the resulting maps significantly higher, highlighting its effectiveness in delivering high-quality and contextually appropriate environments.

The performance evaluation revealed trade-offs between computational performance and map quality, influenced by the structural intricacy of the generated environments. Despite slower performance, the generation times of the proposed method remain acceptable for mobile AR applications. Adapting the method for next-generation XR devices with greater computational power could further enhance its performance and scalability. Techniques such as asynchronous map generation could reduce perceived delays and improve responsiveness, making the method more suitable for real-time or near-real-time applications.

Future work could explore AI-driven techniques to refine generated tiles at runtime, enabling more dynamic and interactive maps. In addition, AI could be utilized to adjust neighboring tiles in response to user modifications, allowing the environment to adapt fluidly and better align with narrative developments. Additional research may also extend the method to support more intricate environments, optimize performance for larger grids, or evaluate its application in broader AR or VR scenarios beyond narrative-driven games. These advancements could establish the method as a versatile tool for procedural content generation in immersive environments.

In conclusion, this work not only addresses the research questions posed at the beginning but also demonstrates the potential of combining RL with procedural generation techniques to meet the unique demands of narrative-driven AR environments. By balancing computational trade-offs with user experience, it lays a strong foundation for future innovations in AR content generation and immersive technologies.

\section*{\uppercase{Acknowledgements}}
I would like to thank Sai Siddartha Maram from the University of California, Santa Cruz for his valuable insights and support during this work.

\bibliographystyle{apalike}
{\small
\bibliography{main}}

\end{document}